# Application of CACS approach for distributed logistic systems

## Sami AL-MAQTARI[1], Habib ABDULRAB[1], Eduard BABKIN[1] [2]

**Abstract.** The article offers original approach which is called Controller Agent for Constraints Satisfaction (CACS). That approach combines multi-agent architecture with constraint solvers in the unified framework which expresses major features of Swarm Intelligence approach and replaces traditional stochastic adaptation of the swarm of the autonomous agents by constraint-driven adaptation. We describe major theoretic, methodological and software engineering principles of composition of constraints and agents in the framework of one multi-agent system, as well as application of our approach for modelling of particular logistic problem.

## 1. INTRODUCTION

Simultaneous rapid grow of logistics market in different regions of the world [1, 2], and its important role in modern economy require wide application of logistics information and management systems for coordinated planning and control. Distributed organizational structure and application of holonic management principles in modern organizations inevitably determine distributed and autonomous features of information systems supporting logistic operations [5]. In such kinds of the systems it is very difficult to apply usual centralized approaches and algorithms for decision support and optimization.

Swarm Intelligence [3, 4] represents one of the interesting paradigm for maintaining self-organization and control in the distributed systems. One of the principal aspect of the swarm-oriented distributed intelligent systems is presence of multiple intellectual and autonomous particles which interact with each other in some way. As it is started in [4]: "Swarm is a population of interacting elements that is able to optimize some global objectives thought collaborative search in space".

Different projects offered approaches for practical application of Swarm Intelligence paradigm in the form of multi-agent systems [6, 28, 30]. Although some of them (i.e. [28]) offer a formal framework for declarative expression and analysis, researchers and practitioners still lack proper generic methods for engineering of the multi-agent systems which have such properties of Swarm Intelligence as emergent behavior, peer-to-peer communication, etc.

Analysis of known logistic problems and algorithms shows that in the domain of applied logistics and optimization general principles of swarm-oriented organization may be realized using proper combination of multi-agent systems (MAS) and constraints satisfaction approach (CSP). So, in this research we pursue the goal to offer a new mechanism of emergent multi-agent behaviour for collaborative search of some feasible solution in accordance with certain inter-agent constraints. In terms of Swarm Intelligence research we replace stochastic adaptation of the swarm of the autonomous agents by constraint-driven adaptation.

In our research we try to satisfy such important requirements of Swarm Intelligence as self-organization and dynamic adaptation to evolving internal or external conditions. Existing approaches to combination of MAS and CSP like [16, 17, 32] do not provide much flexibility and support of dynamic modification of the combined structure of agents and constraints. That's why in this article we propose an original approach which offers a solution for dynamic modification of the combined structure of agents and constraints. Our approach, which was called CACS (Controller Agent for Constraints Satisfaction), allows for joint exploitation of attractive features of the paradigm of multi-agent systems (MAS) and the paradigm of distributed constraint satisfaction (DCSP).

This paper extends and combines our earlier work on joint application of MAS and DCSP paradigms [33, 34]. We describe major theoretic, methodological and software engineering principles of composition of constraints and agents in the framework of one multi-agent system, as well as application of our approach for modelling of particular logistic problem.

The paper is organized as follows. In Section 2 we give background information about MAS and DCSP for better understanding of scientific and technological foundations of our research. In Section 3 we describe main principles of CACS approach. Section 4 contains description of software architecture and implementation principles for software prototype which supports proposed CACS approach. The same section contains overview of used 3d party software platforms. Section 5 describes proposed methodology of practical application of CACS during design and development of DSS. In Section 6 we give overview of the application in ship loading logistics based on CACS prototype. We discuss the achieved results and provide directions for future work in Section 7.

## 2. FOUNDATIONS OF MAS AND DCSP

Paradigm of swarm intelligence is very often and naturally implemented on the basis of multi-agent systems. These systems express major features of collective intelligence [7, 8, 9] and represent the model of problem in terms of autonomous entities that live in a common environment and who share certain resources. The interactions between these individual entities induce cognitive abilities of the whole. Despite multiple-domain-oriented peculiarities majority of multi-agent systems has several significant common features:

---

[1] LITIS Laboratory, INSA Rouen, France. Email: {almaqtari, abdulrab}@insa-rouen.fr.

[2] TAPRADESS Laboraotry, State University – Higher School of Economics, Nizhny Novgorod, Russia. Email: eababkin@hse.ru.

• A limited and local view: every entity has a partial and local knowledge of its environment.

• A set of simple rules: each entity follows a set of simple rules.

• The interactions are manifold: each individual entity has a relationship with one or more other individuals in the group.

•The emerging structure is useful to the community: different entities are a benefit to work (sometimes instinctively) and their performance is better than if they had been alone.

From these points of view, the paradigm of multi-agent systems seek to simulate the coordination of autonomous entities called agents that represent individuals in their community. An agent is an entity that can be viewed as perceiving and acting independently in its environment. According to J. Ferber [10] "One agent called a physical or virtual:

1) which can act in an environment,

2) that can communicate directly with other agents,

3) which is driven by a set of trends (in the form of individual objectives or function of satisfaction and even survival, it seeks to optimize),

4) which has its own resources,

5) which is able to collect (but limited) its environment,

6) which has only a partial representation of this environment (and possibly none),

7) has expertise and provides services,

8) which may be repeated,

9) whose behavior tends to meet its objectives, taking into account the resources and skills available to it and according to its perception, its representations and the communications it receives. "

Given such definition of the agent, we can define a multi-agent system as a set of agents located in a certain environment. They share some common resources, and they interact with each other either directly or indirectly (via their effects on the environment). They seek to achieve the goals of individual agents in the interest of all. The multi-agent systems have applications in the field of artificial intelligence, where they reduce the complexity of solving a problem by dividing the necessary knowledge into sub-units, involving an intelligent agent independent at each of these sub - sets and coordinating the activity of these agents [10].

Because general definitions of inter-agent interaction are too vague we need to apply more strict and formal conventions to express allowable methods of communication between agents. Paradigm of constraints satisfaction, particularly distributed constraints satisfactions, offers flexible and convenient foundations to do this.

The paradigm of constraints satisfaction provides a generic method for declarative description of complex constrained or optimization problems in terms of variables and constraints [12, 13]. Formally, a Constraint Satisfaction Problem (CSP) is a triple $(V, D, C)$ where:

There is $V = \{v_1, \ldots, v_n\}$ is a set of $n$ variables,

a corresponding set $D = \{D(v_1), \ldots, D(v_n)\}$ of $n$ domains from which each variable can take its values from,

and $C = \{c_1, \ldots, c_m\}$ is a set of $m$ constraints over the values of the variables in $V$. Each constraint $c_i = C(V_i)$ is a logical predicate over subset of variables $V_i \subseteq V$ with an arbitrary arity $k : c_i (v_a, \ldots, v_k)$ that maps the Cartesian product $D(v_a) \times \ldots \times D(v_k)$ to $\{0, 1\}$. As usual the value 1 means that the value combination for $v_a, \ldots, v_k$ is allowed, and 0 otherwise.

Constraints involving only two variables are called *binary constraints* [14]. A binary constraint between $x_i$ and $x_j$ can be denoted as $c_{ij}$. Although most of real world problems are represented by non-binary constraints, most of them can be transformed into binary ones using some techniques such as the dual graph method and hidden variable method [15]. Translating non-binary constraints into binary ones allows processing the CSP using efficient techniques adapted only for binary constraints. However, this translation implies normally an increase in number of constraints.

A solution for a CSP is an assignment of values for each variable in $V$ such that all the constraints in $C$ are satisfied. A single solver supports the tasks of collecting all data of the problem: variables, domains and constraints. It treats all such information in a centralized manner.

A Distributed Constraint Satisfaction Problem (DCSP) is a CSP where the variables are distributed among agents in a Multi-Agent System and the agents are connected by relationships that represent constraints. DCSP is a suitable abstraction to solve constrained problems without global control during per—to-peer agent communication and cooperation [16]. A DCSP can be formalized as a combination of $(V, D, C, A, \partial)$ described as follows:

$V, D, C$ are the same as explained for an original CSP,

$A = \{a_1, \ldots, a_p\}$ is a set of $p$ agents,

and $\partial : V \rightarrow A$ is a function used to map each variable $v_j$ to its owner agent $a_i$.

Each variable belongs to only one agent, i.e. $\forall v_1, \ldots, v_k \in V_i \Leftrightarrow V_i \ni \partial (v_1) = \ldots = \partial (v_k)$ where $V_i \subset V$ represents the subset of variables that belong to agent $a_i$. These subsets are distinct, i.e. $V_1 \cap \ldots \cap V_p = \varnothing$ and the union of all subsets represents the set of all variables, i.e. $V_1 \cup \ldots \cup V_p = V$. The distribution of variables among agents divides the set of constraints $C$ into two subsets according to the variables involved in the constraint. The first set is the one of intra-agent constraints $C_{intra}$ that represent the constraints over the variables owned by the same agent $C_{intra} = \{C(V_i) \mid \partial (v_1) = \ldots = \partial (v_k), v_1, \ldots, v_k \in V_i\}$.

The second set is the one of inter-agent constraints $C_{inter}$ that represents the constraints over the variables owned by two or more agents. Obviously, these two subsets are distinct $C_{intra} \cap C_{inter} = \varnothing$ and complementary $C_{intra} \cup C_{inter} = C$.

The variables involved within inter-agent constraints $C_{inter}$ are denoted as *interface variables* $V_{interface}$. Assigning values to a variable in a constraint that belongs to $C_{inter}$ has a direct effect on all the agents which have variables involved in the same constraint. The interface variables should take values before the rest of the variables in the system in order to satisfy the constraints inside $C_{inter}$ firstly. Then, the satisfaction of internal constraints in $C_{intra}$ becomes an internal problem that can be treated separately inside each agent independently of other agents. If the agent cannot find a solution for its intra-agent constraints, it fails and requests another value proposition for its interface variables. To simplify things, we will assume that there are no intra-agent constraints, i.e. $C_{intra} = \varnothing$. Therefore, all variables in $V$ are interface variables $V = V_{interface}$.

Many techniques are used to solve DCSPs. In general the technique proposes a distributed algorithm which is executed by agents that communicate by sending and receiving messages. In general, the messages contain information about assignments of values to variables and rebuttals trust by employees who have no purpose compatible with their own variables. Mainly we mention the Asynchronous Backtracking (ABT) algorithm that was proposed by M. Yokoo [17] and some of its alternatives [18, 19, 20]. These approaches are designed mainly for the treatment of non-binary constraints, however most systems of real constraints are non-binary. Only a few modifications, like [21], were proposed to handle non-binary constraints in the dynamic organization of agents.

## 3. FUSION OF MAS AND DCSP IN CACS APPROACH

In order to avoid shortcomings of known DSCP methods and propose new principles of combination between MAS and DCSP we developed several software engineering methods and algorithms which comprise a new approach for developing DSS. This approach was called Controller Agent for Constraints Satisfaction (CACS). Based on the ABT Algorithm of M. Yokoo [17] CACS approach introduces two types of agents in MAS: Variables' Agent and Controller Agent.

In one hand, a Variables' Agent holds one variable or more. It chooses its values and proposes these values to Controller Agents. On the other hand, Controller Agent encapsulates inter-agents constraints over these variables. Each Controller Agent holds one constraint or more and validates the propositions received from Variables' Agents.

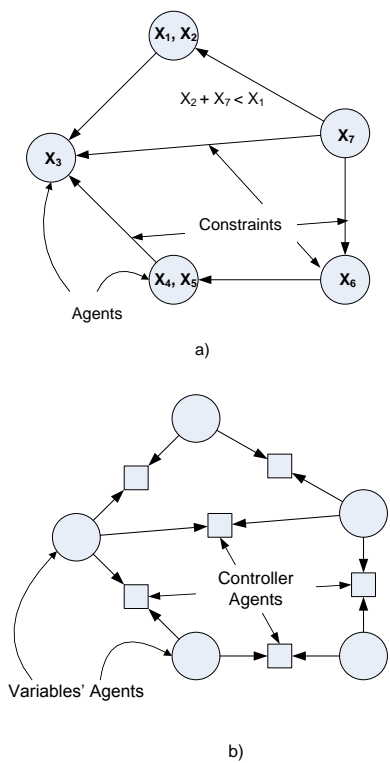

**Figure 1**. A constraint network example: a) without or b) with Controller Agent

We can see in Figure 1 (a) an example of constraint network where Variables' Agent are inter-connected by arcs which represent constraints. These inter-agent constraints are encapsulated in Figure 1 (b) by Controller Agents. The same network can be modified as in Figure 2 by grouping some inter-agent constraints inside a controller agent. With this ability, we can change the scale of constraints grouping from total distribution to total centralization. The problem can vary from designating a controller agent for each constraint to total centralizing by gathering all constraints inside one central controller agents.

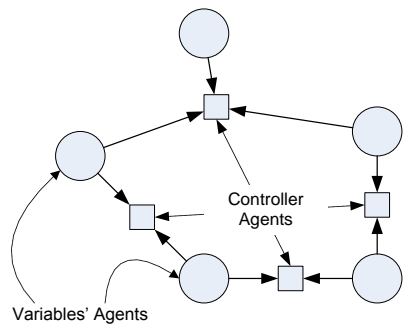

**Figure 2**. Grouping constraints inside Controller Agents.

For abbreviation purposes we will use the term *VAgent* to refer to Variables' Agents and *CAgent* to refer to Controller Agents. In fact, these terms are used as the name of classes used in the implementation of the prototype. The complete DCSP is formulated in terms of VAgents and CAgents. The solution of the problem is seeking during communication between these types of agents. The proposed algorithm of communication is divided into two stages: (1) domain reducing stage and (2) value proposing and validating stage. These stages are explained as follows:

### A. Domain reducing stage

This stage assures constraints consistence by preprocessing variables' domains. The results are reduced domains by eliminating values that would be surly refused by them. This is done as follows:

1. A VAgent sends information concerning the domain of its variable to all linked CAgents. The message takes the form of (variable, domain).
2. After receiving the domains of all variables involved in its constraint, the CAgent uses consistency algorithms [22] in order to reduce these domains to new ones according to its local constraint(s). Then, the controller sends these domains back to their VAgents.
3. Every VAgent receives the new domains sent by CAgents and combines them (by the intersection of received domains) in order to construct a new version of its variable domain.
4. If any new version of a variable domain was empty then we can say that this DCSP is an *over-constrained* problem [23] where no solution can be found. In this case, the system signals that no solution was found (failure). As a prospective, another solution can be investigated by using constraints relaxation [23, 24], in which a VAgent returns to an older version of the domain and reconstruct a new version after neglecting the domains sent by the CAgent that represents the soft constraints that the system may violate according to certain constraint hierarchy [23]. On the other hand, if all variables end with single-value domains then one solution is found. Otherwise, the domain reducing stage is repeated as long as we obtain a different new version of a variable domain. When domain reducing is no longer possible (no more change in variables' domains), we can proceed to the next stage.

The result of the domain reducing stage may be one of the three following kinds: 1) The domain of a variable is reduced to an empty field. Having at least one empty domain for a variable means the problem is over-constrained. If there is no solution that satisfies all the constraints and which contains a value for this variable. 2) The former is reduced to a new domain. This reduction may be the result of responses to a controller or more. This change must be propagated to other controllers. For this, the final stages must be repeated. 3) No change in the domain for this particular variable. In this case, we are faced with two situations: a) there are no changed domains at all. This means that the stage is over and we can proceed with the next stage. b) a change to succeed because of the spread of change in the domain of other variables. These variables can be linked directly or indirectly to the variable concerned.

### A. Value proposing and validating stage

In this stage VAgents make their propositions of values to related CAgents to be tested. Value proposing can be considered as a domain information message in test mode. A test mode means that when a "no-solution" situation occurs because of a proposition the system backtracks to the last state before that proposition. This proceeds as follows:

1. From now on, every VAgent starts instantiating values for its variable according to the new domains. It sends this proposition to the related CAgents.
2. The CAgent chooses the value received from the VAgent with the highest priorities. This value is considered as a domain with a single value. CAgent uses consistency algorithms as in the previous stage to reduce other variables' domains. These new domains are sent to their VAgents to propagate domains change. This step may be viewed as a distributed form of forward checking in an enhanced backtracking algorithm.
3. Like in the previous stage, if all variables end with single-value domains then one solution is found. Unlikely, if the result of this propagation was an empty domain for any variable then the proposed value is rejected and another value is requested. If no more value can be proposed then system signals a no-solution situation to user.
4. If the result of the domain propagation was some new reduced domains with more than one value then steps 1-3 are repeated recursively with the value proposed by the VAgent that have the next priority.

The second stage involves one of three situations: 1) The proposed value is rejected if the spread of this value gives an empty domain for one variable at least. The refusal of a value involves retraction of the former domain and demand for another value. 2) Otherwise, the proposed value is accepted and distributed among the agents. The proposal and validation of values for the other variables continue recursively. 3) If there are more values to be proposed for a variable, the value proposed by the agent who has a higher priority is denied. The algorithm ends in failure when the agent has more priority over proposals valid.

Let's consider an example of MAS where three variables $x$, $y$, $z$ with original permitted domain $\{0, 1, 2\}$ are distributed on three VAgents $A_1$, $A_2$ and $A_3$, and two constraints exist: $x \neq y$ and $x + y < z$. These constraints are placed into two CAgents $C_1$ and $C_2$.

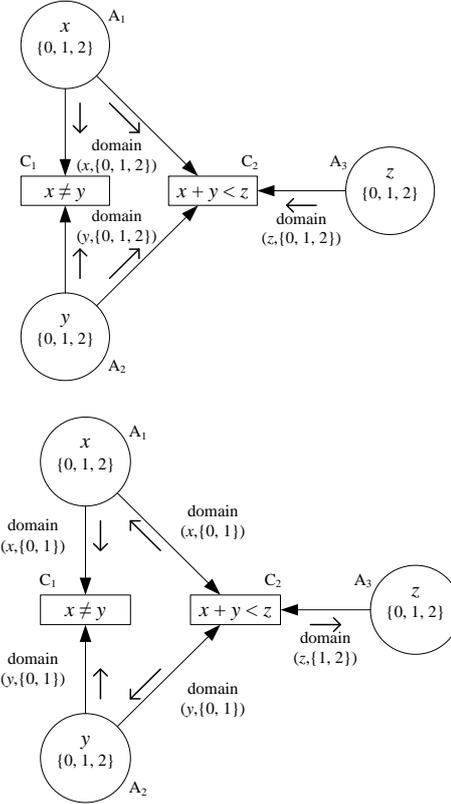

**Figure 3.** Illustration of domain reducing stage of CACS.

During the first stage of CACS (fig.3) three agents $A_1$, $A_2$ and $A_3$ are sending the domain $\{0, 1, 2\}$ for the three variables $x$, $y$ and $z$ respectively agents $C_1$ and $C_2$. $C_1$ tries to reduce the domains of $x$ and $y$. Obviously, no change is possible. On the contrary, the agent $C_2$ changes the domains of variables $x$ and $y$ in $\{0, 1\}$ and the domain of $z$ in $\{1, 2\}$. This change will be propagated to the agent $C_1$ which returns the same domains for variables $x$ and $y$ (i.e. $\{0, 1\}$). The domain reducing stage finishes with the domain $\{0, 1\}$ for the variables $x$ and $y$ and the $\{1, 2\}$ for the variable $z$.

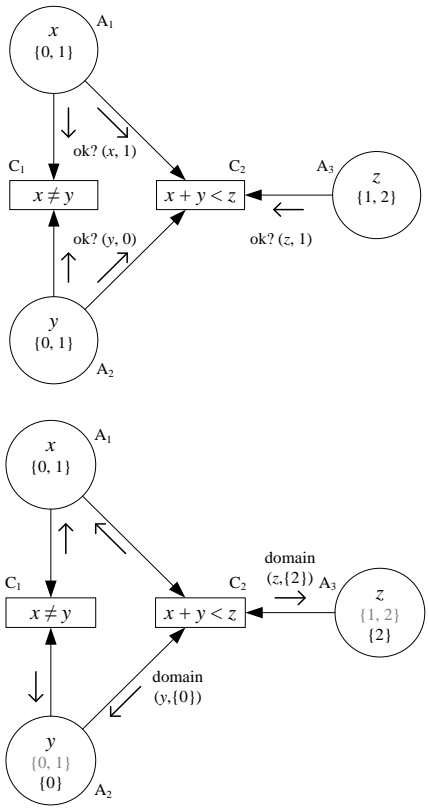

**Figure 4**. Illustration of value proposing and validating stage of CACS.

During the second stage (fig.4), algorithm will assign priorities to the agents A1, A2 and A3 according to their index. So the agent A1 will have the highest priority, and the agent A3 will have the lowest priority. Suppose that the agent A1 proposes value 0 for the variable x to the agents C1 and C2. C1 treats this value as the domain {0} and reduces the domain of the variable y to {1}. The spread of this new domain reduces the domain of the variable z to {2}. A2 tries to offer as the value 0 for variable y. His proposal will be of lower priority than the agent A1 and will be refused because they are inconsistent. The same result is obtained for any other value.

According to the results of the first and second stages, we can say that the CACS algorithm solves DCSP: 1) When the DCSP is over-constrained, we are faced with two different situations: Either the initial domains of the variables are inconsistent. This means that at the end of the first stage there is at least one empty domain of a variable. This involves termination of the algorithm and the declaration of a state of non-solution. Either the initial domains of the variables are consistent. 2) Where there is a unique solution of DCSP, we face two situations: The domains are consistent as long as there is a solution to the DCSP. If the first stage ends with single-vale domains, it means that the solution is found and the algorithm stops. Otherwise, in the second stage, the value proposed by a variable if it is not inconsistent with a value proposed by another agent with higher priority. The proposals of the agent with the highest priority are a priori accepted by all CAgents (it is necessary that this value is part of the final solution to be finally accepted). 3) When the DCSP is under-constrained, many solutions exist. The order of each proposed agent determines convergence towards any particular solution. In other words, the agents start the proposals by the most suitable for their purposes. For example, if an agent tries to minimize the value of its variable, it must begin proposing values from the minimum to the highest values.

## 4. SOFTWARE IMPLEMENTATION OF CACS

To prove the proposed methods of constraints satisfaction based on two types of the agents we developed an object-oriented CACS software prototype which can be considered as a generic framework for distributed information syste4ms in logistics. As we can see from Figure 5, the developed CACS prototype uses hierarchical multiple-layer architecture.

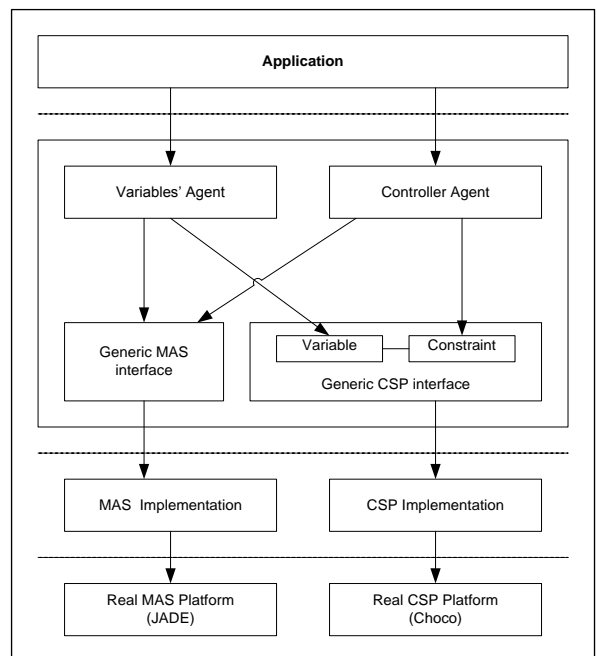

**Figure 5**. Software architecture of CACS prototype.

This architecture allows developing applications more flexibly by separating it into specialized layers. The very top layer is the application layer which is the implementation of a DCSP problem using the proposed system underneath it. From the application view point, the system is composed directly from the two principal types of agents: the CAgent and the VAgent. Both agents are inherited from CommonAgent class that defines some shared functionalities between both types of agents. The user can create the necessary VAgents according to its problem definition. He also creates the constraints and associates them to CAgents.

The second layer is the intended system (CACS) where our two-stage interaction algorithm is implemented in accordance with previous definition. Figure 7 shows the interaction between agents during the domain reducing stage. The interaction protocol is a loop of repeated domain informing from the VAgents side to CAgents side and new domain proposing as response. This loop is repeated until no further domain reduction is possible (or an empty reduced domain is found which signify that there is no solution).

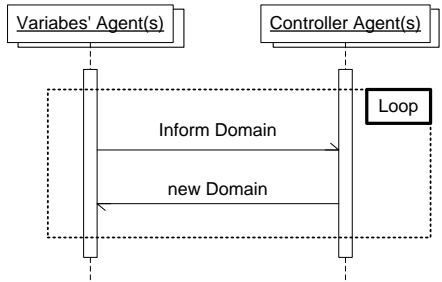

**Figure 6**. Implementation of interaction during domain reducing stage

The interaction between agents during value proposing stage is shown in Figure 6. as nest loops: the internal loop is similar to the domain reducing loop in Figure 6. Variables' domains are reduced according to the proposed value in the external loop. In the external loop, values are proposed and evaluated after the domain reduction to be either accepted or rejected. The external loop continues until we obtain single value domains for all variables.

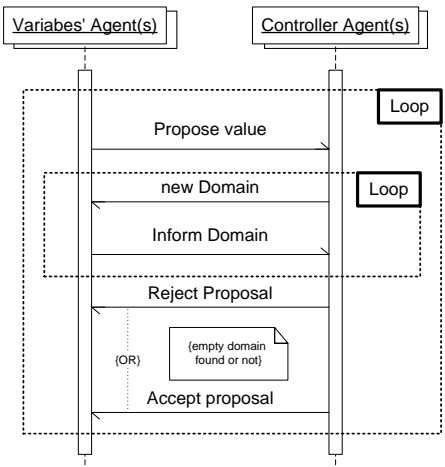

**Figure 7**. Implementation of interaction during value proposing stage

The system layer uses generic interfaces for both MAS and CSP platforms. This allows the system to use any existing MAS and CSP platforms by implementing these interfaces. At the same time this isolates the internal structure from the changes of choice of platforms. An intermediate layer between the system and the real MAS or CSP platform is necessary in order to separate the structure of the system from that of the real MAS and CSP platforms. This layer works as an adapter; it implements the generic platforms in the system layer using the real platforms. This implementation difficulty varies according to the MAS and CSP platforms used for the realization of the final system.

The whole CACS prototype was developed in Java language. Due to the object oriented nature of Java language agents and the messages are represented by objects (Figure 8, 9).

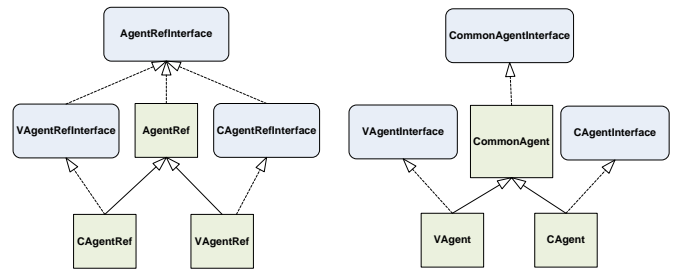

**Figure 8**. The hierarchy of the main components of agents (agents and reference to the agents). Rectangles with rounded corners represent interfaces; rectangles with sharp corners represent classes

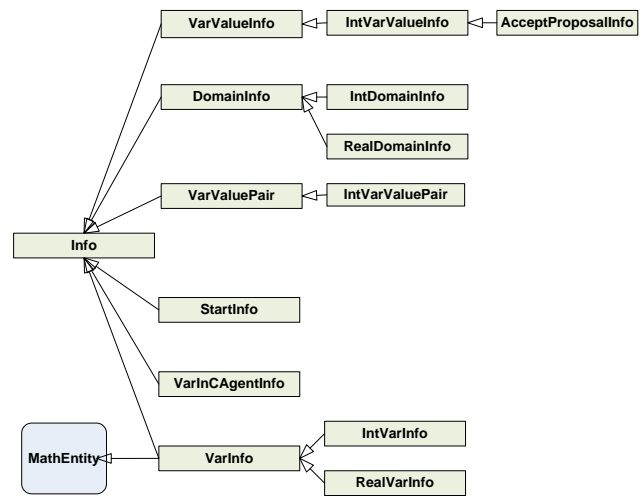

**Figure 9.** The hierarchy of agent messages. Rectangles with rounded corners represent interfaces; rectangles with sharp corners represent classes

However, from the point of view of Multi-Agent System design, agents should not be referenced by a simple public reference that is accessible by any other object in the system. The reason for that is to prevent any direct access to the agent internal functionality. Normally, references to agents should be kept hidden by the MAS platform and communicating with an agent is made by messages that would be delivered by the system using the agent address. Mapping from agent address to its real reference is an internal functionality of the MAS platform.

In order to be more generic, we distinguish in the prototype implementation between the agent and its reference. For this purpose, VAgentRef and CAgentRef classes have been designed. Both classes are inherited from the abstract AgentRef class. They are used as references to either variables' agents or controller agents. When an instance of the class DCSP is used to create an instance of VAgent or a CAgent, it returns an instance of either VAgentRef or CAagentRef classes respectively according to created agent. In the same manner, a variable inside variables' agents cannot be referred directly. In fact, a controller agent keeps a copy of that variable inside it and propagates any change on that variable to the owner agent. Instead of dealing with variables directly between agents, they deal with variables identifiers. A variables identifier is an instance of VID class. It is simply the name of the variables and the identifier of its owner agent. An instance of VAgentRef is used to create variables inside the corresponding VAgent. A variable creation process returns an instance of VID class identifying the created variable.

Among additional features we added to our prototype a possibility to declaratively define a simple DCSP via the use of XML notation. The XML file that describes a DCSP problem should be built according to the following model (fig.10):

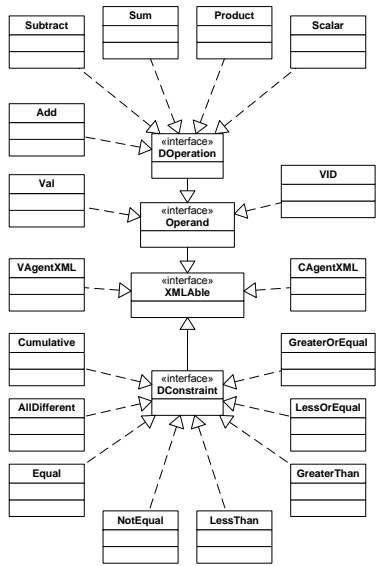

**Figure 10.** The hierarchy of the main components of agents (agents and reference to the agents)

The choice of multi-agent platforms and multi-solver constraints required a study and testing of several platforms. We reviewed our work over multiple platforms including JADE and Madkit and several constraints solvers as CHOCO, Cream and JCK. Finally we chose for the role of MAS JADE (Java Agent DEvelopment Framework) multi-agent framework [25], and for CSP platform, we have chosen Choco [26, 28].

JADE is a multi-agent framework compliant with the FIPA specifications [27] and is fully implemented in Java language. JADE was established by the laboratory TILAB Telecom Italia. JADE has three main modules (fig.11): DF (Directory Facilitator): provides a service of "yellow pages" to the platform; ACC (Agent Communication Channel) handles communication between agents; AMS (Agent Management System) oversees the registration of agents, authentication, access and use of the system. Each JADE agent is composed of a single thread of execution (thread). Each task agent is represented by an instance of class Behavior. Jade offers the possibility of agents' multi-threaded, although the user leaves the responsibility for managing competition (except the timing of the messages file ACLs).

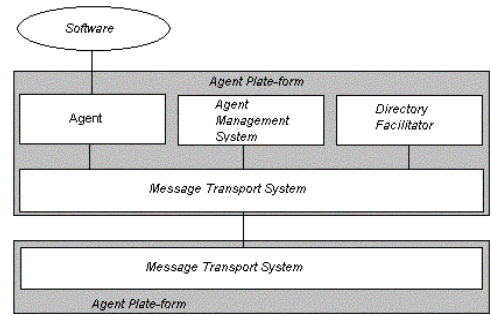

**Figure 11.** Architecture II software platform JADE

In order to implement a behavior, the developer must define one or more objects of class Behavior, the instantiate and add them to the thread of execution of the agent. Every object type has a Behavior method *action ()* (which is the treatment to be performed by it) and a method d*one ()* (which checks if the treatment is completed). In detail, the scheduler executes the method *action ()* of each object in the queue of the tasks of the agent. Once this is completed, the method *done ()* is invoked. If the task has been completed then the Behavior object is removed from the queue. The scheduler is non-preemptive and does only one behavior at a time, one can consider the method *action ()* as atomic. It is then necessary to take certain precautions during the implementation of the latter, to avoid endless loops or operations too long. The most classic program behavior is to describe it as a finite state machine. The current status of the agent is stored in local variables.

Also JADE simplifies the implementation of multi-agent systems through a set of graphical tools that supports the debugging and deployment phases.

Choco is a library for constraint satisfaction problems (CSP), constraint programming (CP) and explanation-based constraint solving (e-CP) [28]. It is built on an event-based propagation mechanism with backtrackable structures. Choco is implemented in Java and takes advantage of the principle of inheritance to allow the programmer to define its own types and constraints. This is achieved by using abstract classes (fig. 12):

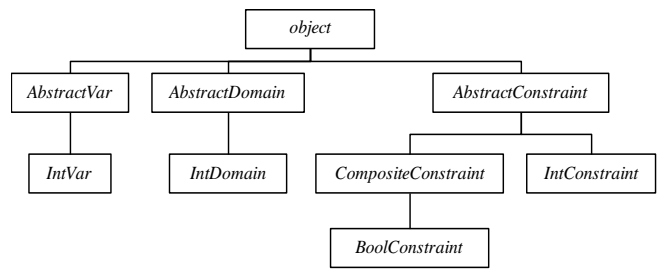

**Figure 12.** Hierarchy of constraints in Choco

It permits the use of multiple solvers for different problems separately. This allows each CAgent to have its own solver. A distributed constraint problem is created as an instance of the class DCSP. This instance represents the problem to be solved and is used to create the different needed agents.

Our prototype in its current state is composed of three main packages containing more than 80 classes and Java interfaces and approximately 4300 lines of code.

## 5. DESIGN METHODOLOGY IN CACS

A specific methodology was designed to allow the user to develop distributed multi-agenty systems using Swarm Intelligence paradigm and CACS approach. In general this methodology consists of the following steps:

1. Identify the key actors of the problem (VAgents). These actors are the entities of the system modeled.
2. Determine the properties (variables) of these actors that are restricted by constraints with properties of other actors.
3. Determine all the constraints of the problem.
4. Classify constraints logically in separate groups.
5. Specify a set of Controller Agents to monitor each group of constraints.

To provide a developer with flexible practical methods of the design we offer two refinements of the general methodology: simple and complex.

To prove the proposed methods of constraints satisfaction based on two types of the agents we developed an object-oriented CACS software prototype which can be considered as a generic framework for distributed model-driven DSSs. As we can see from Figure 5, the developed CACS prototype uses hierarchical multiple-layer architecture. The following steps correspond to a given DCSP:

6. Creation of the problem P. This is done by creating an instance of the class DCSP from the package dcsp.
7. Creation of agents to control variables (specifically, their references) via the prolem P. using the method *makeVAgent* () to create a variable and method *makeCAgent* () to create a controller. 3) Creating variables distributed via agents which own variables. This is done through the method *makeBoundedIntVar* () which creates a variable with two upper and lower limits.
8. Creation of constraints on variables.
9. Addition of constraints to CAgents.
10. Start the algorithm of resolution through the DCSP P.

The use of the prototype can be demonstrated via the following simple example:
$V = \{x, y, z\}$ is the set of variables from the domain $\{1, \ldots, 100\}$ for all of them, $C = \{c_1, c_2, c_3\}$ is the set of constraints:

$$c_1 : x \neq y, y \neq z, x \neq z \text{ (or alldifferent } (x, y, z))$$
$$c_2 : x \geq y$$
$$c_3 : z \geq y$$

In order to model this problem using the proposed prototype the user should proceed as follows. We start by assigning variables to VAgents. In this example, agents v1, v2, and v3 own variables x, y, and z respectively. Note that the distribution of variables may be a problem dependant issue which means that the user chooses the owner agent of each variable according to the problem specifications. In the same manner, constraints also should be assigned to CAgents. In this example, we assign each constraint to a CAgent.

1. Create a distributed problem p (an instance of DCSP class). This class will be used in order to create VAgents and CAgents and to start our CACS algorithm.
```
DCSP p = new DCSP("example");
```
This creates a distributed problem with which agents, variables and constraints will be created.

2. Use this instance to create both types of agents. This is done by calling makeVAgent() and makeCAgent() methods from the DCSP instance created in step 1 as follows:
```
VAgentRef v1 = p.makeVAgent ("v1");
VAgentRef v2 = p.makeVAgent ("v2");
VAgentRef v3 = p.makeVAgent ("v3");
CAgentRef c1 = p.makeCAgent ("c1");
CAgentRef c2 = p.makeCAgent ("c2");
CAgentRef c3 = p.makeCAgent ("c3");
```

3. Create variables inside VAgents. In other word, assign variables to variables agents. The method makeBoundedIntVar() is used to achieve this as follows:
```
VID x = v1.makeBoundedIntVar ("x", 1, 100);
VID y = v2.makeBoundedIntVar ("y", 1, 100);
VID z = v3.makeBoundedIntVar ("z", 1, 100);
```

4. Create the constraints and post them to CAgents. The constraints are created separately and posted to their owner agents using the method post():
```
c1.post(new AllDifferent(new VID[]{x,y,z}));
c2.post(new GreaterOrEqual(x, y));
c3.post(new GreaterOrEqual (y, z));
```

5. Start the CACS algorithm by calling solve() method from the DCSP instance:
```
p.solve();
```
This last instruction initiates communication between the different agents in the system in accordance the algorithm described previously in Section 3. If an agent finds a value for its variable that corresponds to a solution then it will notify to this value. The solution will be the combination of all values from all agents. Otherwise, no-solution state is declared.

Also the developer can express the structure of DCSP in declarative manner using XML. For instance, the problem described in previous sub-section can be written in XML as follows:
```
<?xml version="1.0" encoding="UTF-8"?>
<!DOCTYPE dcsp SYSTEM "dcsp.dtd">
<dcsp>
<name>example</name>
<vagent><name>v1</name><var><name>x</name>
    <inf>1</inf><sup>100</sup></var></vagent>
v2 and v3 by the same manner
<cagent>
    <name>c1</name>
    <constraint><alldiff>

<vid><name>x</name><owner>v1</owner></vid>

<vid><name>y</name><owner>v2</owner></vid>

<vid><name>z</name><owner>v3</owner></vid>
    </alldiff></constraint></cagent>
c2 and c3 by the same manner
</dcsp>
```

# 6. CACS APPROACH IN TRANSPORT LOGISTICS

We consider modern transportation problems as a natural candidate domain for evaluation of the proposed CACS approach. Although there is a lot of different centralized algorithms in this area we believe that multi-agent techniques can radically improve efficiency and fairness of negotiation between participants in the course of problem solving as well as improve reactivity of the logistics systems. Among different benefits of logistics management within the CACS framework we can point out such positive features as: better consideration of individual preferences and ability of their dynamical changes in the course of solving, early availability of partial solutions and inherently distributed structure of the system.

In order to create solid foundations for application of Swarm Intelligence and CACS approach in transportation logistics we developed a distributed multi-agent application which mimics major features of modern ship loading problems, and evaluated its feasibility and performance. Our CACS application is based on a simplified ship loading scenario which was originally presented in studied in Chips constraint solver by Kay Chips (Kay 1997) and later was expressed in terms of Java-based Choco constraint solver by prof. A. Aggoun.

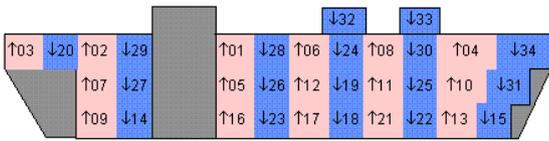

**Figure 13.** Graphical representation of the original Kay's ship loading problem (Kay 1997)

In the discussed problem a specific precedence function *pred* in defined over the loading items. For each of the items the number of workers needed for loading is specified.

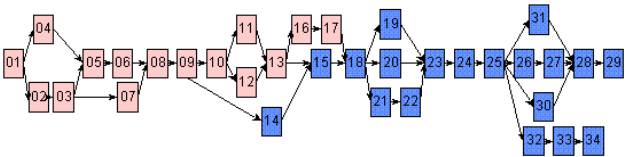

**Figure 14.** The feasible order of loading tasks (the loading plan) in accordance with the constraints given (Kay 1997)

According to CACS methodology each loading task is realized as a separate Variable Agent in our CACS application. Variable Agent holds three specific variables. These variables determine start time of loading ($t^i_{start}$), finish time of loading ($t^i_{end}$) and predetermined loading duration ($d^i$) accordingly.

All constraints of the considered problem are grouped inside Controller Agents. We recognize three different groups of Controller Agents according to the semantics of the constraints. The first group contains Controller Agents which hold duration constraints. The agent of that group is responsible for verifying that the loading tasks are scheduled within the time frame. It means that for each task $i$ the following constraint should be satisfied: $t^i_{start} + d^i \leq t^i_{end}$.

The second group contains Controller Agents which are responsible for verifying that the loading plan satisfies precedence constraints given (like one on the fig. 13). Finally the third group contains Controller Agents which are responsible for verifying availability of the resources for the loading plan. Controller Agent of that kind holds cumulative constraint over the number of workers available for finishing the ship loading within the total time.

That cumulative constraint may be expressed using current values of $t^i_{start}$ and $t^i_{end}$ variables, as well predetermined workforce effort needed for each task $w^i$. Given these values we can define the scheduling matrix $SC$.

$$SC = \begin{bmatrix} 0 & 0 & w^1 & w^1 & w^1 & 0 & 0 & ... & 0 \\ 0 & 0 & 0 & w^2 & w^2 & 0 & 0 & ... & 0 \\ & & & & & & & & \\ 0 & 0 & 0 & 0 & 0 & 0 & 0 & ... & w^N & w^N \end{bmatrix}$$

The element $SC_{ij}$ is equal to $w_i$ iff at the time moment $j$ the loading task $i$ is performed, and it is equal to 0 in the opposite case.

Using that matrix we may define the maximum number of workers needed at each moment of the time and the needed cumulative constraint:

$$\max_j \left( \sum_{i=1}^{N} SC_{ij} \right) \leq \max Persons.$$

With such problem interpretation we may completely describe it in terms of our CACS approach. The original structure of the agents is presented on Figure 15.

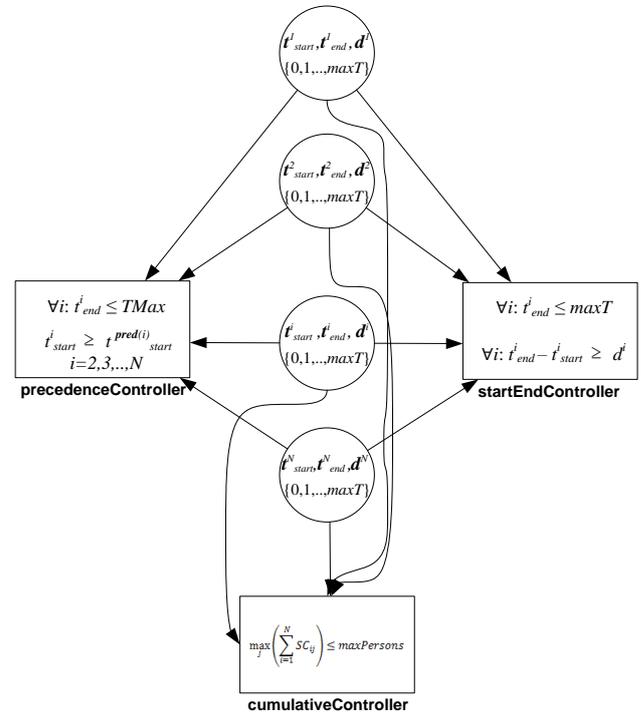

**Figure 15.** Connected structure of Variable Agents (circle) and Constraint Agents (rectangle) for the ship loading problem

Using the proposed methodology we designed the Java-based application that solves the ship loading problem. In that program at the first moment the ControllerAgents are created:

```
CAgentRef startEndController =
  dpb.makeCAgent("startEndController");
CAgentRef precendenceController =
  dpb.makeCAgent("precendenceController");
CAgentRef cumulativeController =
  dpb.makeCAgent("cumulativeController");
```

Then auxiliary VariableAgent is created which stores the total time of loading operations:

```
VAgentRef general = dpb.makeVAgent("General");
VID generalEnd = general.createVar("General_End",
  0, timeHorizon);
```

After that in the cycle thirty-four VariableAgents are created which correspond to the loading tasks and store needed variables $t^i_{start}$, $t^i_{end}$ and duration $d^i$. In the same cycle the duration constraints are created and attached to the corresponding ControllerAgent.

```
for (int j = 0; j < nbTasks; j++) {
taskAgents[j] = dpb.makeVAgent("task_agent_" +
  (j + 1));
taskStarts[j] = taskAgents[j].
  createVar("Start", 0, timeHorizon);
taskEnds[j] = taskAgents[j].createVar("End",
  0, timeHorizon);
taskDurations[j] = taskAgents[j].
  createVar("Duration", durations[j],
  durations[j]);

DOperation startEndOperation = new
  Subtract(taskStarts[j], taskEnds[j]);
DConstraint startEndConstraint = new
  Equal(startEndOperation, taskDurations[j]);
startEndController.post(startEndConstraint);

DConstraint endConstraint =
  new LessOrEqual(taskEnds[j], generalEnd);
startEndController.post(endConstraint);
}
```

Finally the precedence constraint and cumulative constraint are determed and the process of solution search is started.

With the given conditions in the result of application run the values of variables $t^i_{start}$, $t^i_{end}$ will constitute a feasible solution ot the ship loading problem.

## 7. CONCLUSIONS & FUTURE WORK

In this article we proposed a new approach for combination of MAS and DCSP in multi-agent swarm systems. This approach called CACS (Controller-Agent for Solving Constraints) based on the use of a specific type of agents called Agent Controller and Variables' Agent. We believe that proposed process of constraint satisfaction in the multi-agent system fits well the general principles of Swarm Intelligence. In particular, the stage of domain reduction in our algorithm may be seen exchange of "rules, tips and believes about how to process the information [4]".

Also in our approach we implemented a principal feature of Swarm systems, which is principal ability to modify multi-agent structure in response of various influencing factors. First of all, declarative manner of constraints based formalization of the problem allows for changing inter- and intra-agent behavior. Secondly, the composition of inter-agent constraints inside ControllerAgent may be changed during evolution of the system (as it shown on fig.1 and fig .2).

In the proposed CACS architecture we see good opportunities for further moving towards to implementation of advanced swarm intelligence capabilities. Modern MAS platforms like JADE implement different peer-to-peer communication mechanisms for which direct correspondence may be found in computational biology. Given such mechanisms as foundation for reliable distributed inter-agent communication we will extend discussed algorithms of interaction between Controllers Agents and Variables Agents by adaptation framework. In such framework agents will be able to discover critical changes in MAS configuration (faults of agents, misbehavior, etc), negotiate responsibilities and change the roles accordingly in order to continue proper collective operations.

The model of distributed constraints satisfaction proposed in SACS also offers two main contributions in DCSP research. First, it is the possibility of a direct and easier dealing with non-binary constraints without having to use methods of transformation of non-binary constraints to binary constraints. Second, CACS offers us the possibility to organize the constraints logically related groups. This grouping of constraints allows us to form sub-problems, each group is monitored and processed by a single controller. This also helps reduce the total number of Controller Agents needed.

Non-binary constraints are more common in real problems than binary ones. Some methods are used in order to allow using binary constraint solving techniques on non-binary ones. Methods like hidden and dual transformation [14, 15] convert non-binary constraints into equivalent binary ones. Other methods are proposed in the DCSP domain in order to deal with non-binary constraints. I. Brito [21, 33,34] has proposed organizing agents involved in a non-binary dynamically in order to form a proper precedence-validate sequence. Agents then follow that sequence to find a solution for that constraint.

Our algorithm proposes another direct alternative. Any constraint is encapsulated inside a controller agent regardless this constraint is binary or non-binary. Agents involved in any constraint are not forced to follow any order in proposing values for their variables.

The increase in number of agents is an inconvenience of our model. We can investigate the possibility of using a hybrid system of both, our model and a standard ABT model, in order to model a DCSP. In such hybrid system, binary constraints relate variables' agents directly while non-binary constraints are encapsulated inside controller agents. The possibility of gathering constraints gives also the possibility of decreasing the number of agents. The user can group some constraints according to the modeled problem logic.

To prove the feasibility of the proposed theoretical principles we implemented software prototype of CACS. It uses generic interfaces for integration with different third-party MAS-platforms and CSP-solvers. In the final implementation we used the MAS platform JADE and the Choco CSP solver. Apart from direct Java programming of DCSP problems our prototype also provides an opportunity to describe the problem using XML facilitating the modeling of simple problems without the need to write and compile a Java program.

Demonstrated applicability of CACS for solution of logistics problems opens opportunity for further progress in developing Swarm Intelligence applications. Following that direction we plan to continue in design of meta-communication protocol between ControllerAgents, which will permit define formal methods of re-composition of constraints inside different ControllerAgents during evolution of the system.

Another interesting problem for CACS application comes from the domain of modern transportation systems. Here we wish to apply CACS approach for the "transport on demand" challenge and solution of complex logistics problems in real conditions of modern warehouses. Also we are going to investigate ways to add optimization mechanism to the system similar to DPOP algorithm [34]. This will allow the user to adjust the Variables' Agent value choosing according to a given optimizing mechanism.

This work was partially supported by HSE grant # T3-61.1.